\documentclass{article}
\usepackage{spconf,amsmath,graphicx}
\usepackage{url}
\usepackage{enumitem}
\setlist{nosep, leftmargin=14pt}

\makeatletter
\def\@name{ \emph{Wenhui Cui$^{1}$, Woojae Jeong$^{1}$, Philipp Thölke$^{2}$, Takfarinas Medani$^{1}$,}  \\ \emph{Karim Jerbi$^{2, 3, 4}$, Anand A. Joshi$^{1}$, Richard M. Leahy$^{1}$}}
\makeatother

\usepackage{mwe} % to get dummy images

\usepackage{multirow}
\usepackage{booktabs}
\title{Neuro-GPT: Towards A Foundation Model for EEG}
\address{\\ $^{1}$ Ming Hsieh Department of Electrical and Computer Engineering, University of Southern California, \\ Los Angeles, CA, USA \\ $^{2}$  Psychology Department, Université de Montréal, Montreal, QC, Canada  \\$^{3}$ Mila (Quebec AI research institute), Montreal, QC, Canada, \\$^{4}$ UNIQUE (Quebec Neuro-AI research center), QC, Canada}

\begin{document}
%\ninept
%
\maketitle

\begin{abstract}
To handle the scarcity and heterogeneity of electroencephalography (EEG) data for Brain-Computer Interface (BCI) tasks, and to harness the power of large publicly available data sets, we propose Neuro-GPT, a foundation model consisting of an EEG encoder and a GPT model. The foundation model is pre-trained on a large-scale data set using a self-supervised task that learns how to reconstruct masked EEG segments. We then fine-tune the model on a motor imagery classification task to validate its performance in a low-data regime (9 subjects). Our experiments demonstrate that applying a foundation model can significantly improve classification performance compared to a model trained from scratch, which provides evidence for the generalizability of the foundation model and its ability to address challenges of data scarcity and heterogeneity in EEG. The code is publicly available at \url{https://github.com/wenhui0206/NeuroGPT}.
\end{abstract}

    
\section{Introduction}
The limited scale of training data for electroencephalography (EEG) based Brain-Computer Interface (BCI) classification tasks poses challenges to applying deep learning models.
These models require a large amount of training data to converge and generalize to unseen testing data. However, individual differences can lead to heterogeneous feature representations across subjects~\cite{du2022eeg}, which makes it difficult to generalize the model across subjects. EEG's high-dimensional nature and limited availability for specific tasks create additional barriers to the convergence of these models. 

% One of the most straightforward solutions is transfer learning, in which knowledge from large-scale datasets is leveraged to benefit the task of interest, provided the two domains have some information in common~\cite{azizi2021big}. 
One common approach is to learn generalizable features from large amounts of data using self-supervised learning and then transfer to the task of interest~\cite{reed2022self}. Here, we address the question of whether we can train a model on large-scale EEG datasets using a self-supervised task and then transfer the pre-trained knowledge to enhance performance on a downstream task. Large language models (LLMs) in natural language processing (NLP) tasks have proven extraordinarily successful using this approach. Similar models have also shown remarkable performance in other tasks including image and video generation~\cite{yu2023language}, medical question answering~\cite{singhal2023expertlevel}, and neural activity analysis~\cite{thomas2022self, ortega2023brainlm, azabou2023unified}. 
Despite the popularity of LLMs, there have been relatively few attempts to adapt them to EEG data. 
The work in~\cite{kostas2021bendr} employed a BERT-inspired~\cite{devlin2018bert} approach to pre-train a transformer model on massive EEG data using a contrastive self-supervised task. However, it exhibited limited generalizability to downstream tasks. 
%Here we pre-train an EEG foundation model on large-scale public EEG datasets and then fine-tune the model on downstream BCI tasks, where data is scarce and heterogeneous. 
% Then we input a sequence of contiguous chunks into the GPT model, which then learns to predict the next masked chunk (token). This process enables the model to understand the temporal dependencies and patterns within the EEG data. However, employing chunks of raw EEG signal as input tokens to GPT model is problematic. Given the high-dimensionality and low signal-to-noise ratio of EEG data~\cite{bang2021spatio, lai2018artifacts}, predicting raw signals is particularly challenging for the GPT model and it may not learn meaningful features given the presence of noise. Moreover, the GPT model only captures temporal dependencies while EEG data exhibits rich spatio-temporal dynamics~\cite{du2022eeg}. Therefore, we introduce an EEG encoder~\cite{song2023eeg} comprising convolutional and transformer layers, to extract spatio-temporal features from raw EEG signals. The learned embeddings serve as a lower-dimensional and denoised representation of the raw EEG signals, not only simplifying the prediction of the masked chunk for the GPT model but also enhance its ability to capture informative temporal correlations and patterns.

Here we aim to lay the groundwork for developing foundation models for EEG. We employ a 
Generative Pre-trained Transformer (GPT) model~\cite{radford2019language}, which uses a decoder-only transformer architecture and is trained to predict the next masked token given a sequence of tokens as input (auto-regressive training). In text-based tasks, a sentence is broken down into tokens as input units to the model. A token can be a few characters or a word, depending on the language and context.
To adapt the GPT model to EEG data, we split the whole time series into fixed-length ``chunks", treating each chunk as a token. An EEG encoder is incorporated to extract representative features from raw EEG data.
Neuro-GPT is a foundation model consisting of an EEG encoder to extract spatio-temporal features from EEG data, and a GPT model that uses self-supervision to predict the masked chunks. The foundation model is pre-trained on the TUH EEG dataset~\cite{obeid2016temple}. We fine-tune the model on a motor imagery classification task where only 9 subjects are available. Experiments showed that the EEG encoder learns meaningful features that are generalizable to the downstream task.
%, and the GPT model serves as an auxiliary component to perform the self-supervised task and assist the EEG encoder in extracting expressive features from raw EEG during pre-training. 
%The EEG encoder is then adapted to generate the most effective features for classification during fine-tuning. 
% We experimented with various input configurations and model hyper-parameters to find the optimal pre-training setting
%ALREADY IN ABSTRACT AND IN DIFFERENT WORDS IN THE DISCUSSION: Experiments demonstrated that applying the proposed foundation model can significantly improve classification performance compared to the model trained from scratch, which provides evidence for the advanced generalizability of foundation model and overcoming the challenges of data scarcity and heterogeneity.

\section{Methods}
\begin{figure}[htb]
\begin{minipage}[b]{1.0\linewidth}
  \centering
\centerline{\includegraphics[width=8.5cm]{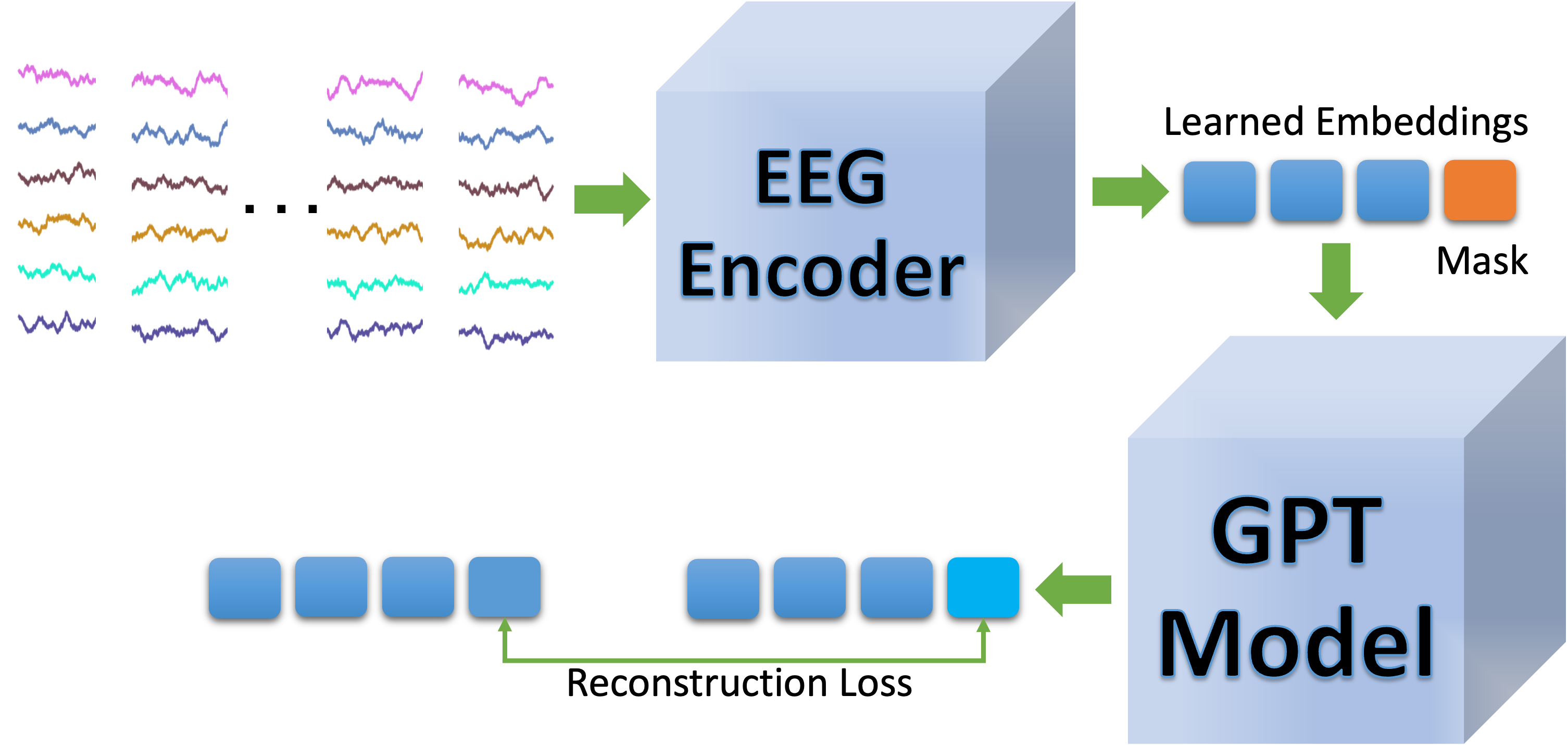}}
%  \vspace{2.0cm}
  % \centerline{Neuro-GPT}\medskip
\end{minipage}
\caption{Neuro-GPT Pipeline: the EEG encoder takes chunks of EEG data as input and generates embeddings as tokens for the GPT model. The last embedded chunk in the sequence is masked. The GPT model then predicts the masked chunk and a reconstruction loss is computed between the prediction and the original embedding token.}
\label{fig:pipeline}
\end{figure}

In this section we introduce the architecture of Neuro-GPT, the pre-training details, and the fine-tuning strategies.
We divide the raw EEG data into fixed-length chunks from which we generate a sequence of tokens corresponding to contiguous data chunks. The GPT model then learns to predict masked tokens. Employing chunks of raw EEG signals directly as input tokens to the GPT model would be problematic. Given the high dimensionality and low signal-to-noise ratio of EEG data~\cite{lai2018artifacts}, predicting raw signals is particularly challenging for the GPT model, and it may not learn meaningful features given the presence of noise. 
% Moreover, the GPT model only captures temporal dependencies, while EEG data exhibits rich spatio-temporal dynamics~\cite{du2022eeg}. 
Thus, we introduce an EEG encoder~\cite{song2023eeg} comprising convolutional and transformer layers to extract spatio-temporal features from the raw EEG. We input chunks of EEG into the encoder to generate the embeddings. These embeddings serve as a lower-dimensional and denoised representation of the raw EEG signals, not only simplifying the prediction of the masked chunk for the GPT model but also enhancing its ability to capture informative temporal correlations and patterns.
The overall Neuro-GPT pipeline is illustrated in Figure~\ref{fig:pipeline}.
\subsection{Neuro-GPT Pipeline}
\label{sec:causal}
\textbf{EEG Encoder }
We adopt an encoder architecture incorporating both convolutional and self-attention modules. This arrangement has achieved state-of-the-art performance in BCI classification tasks~\cite{song2023eeg}. We split the raw EEG signals into $N$ chunks, each of time length $T$. This results in a sequence of chunks denoted $\{D_1, D_2, \cdots, D_N\}$. Each chunk is of dimension $C\times T$, where $C$ is the number of channels. Each chunk is treated as an individual training sample in the encoder. In the convolutional module, we apply a temporal convolution filter to the time series and a spatial convolution filter to the electrodes of the EEG. Then after average pooling, the extracted local features are fed into the self-attention layers to incorporate temporal dependencies within a chunk.
The self-attention mechanism combined with convolution will encode the spatio-temporal features of each chunk. 
The outputs of the encoder are the embedded chunks or tokens: $\{\mathcal{H}(D_1), \mathcal{H}(D_2), \cdots, \mathcal{H}(D_N)\}$, where $\mathcal{H}$ denotes the mapping learned by the EEG encoder from raw EEG signals to embeddings.

\begin{figure}
  \centering
 \includegraphics[width=5.0cm]{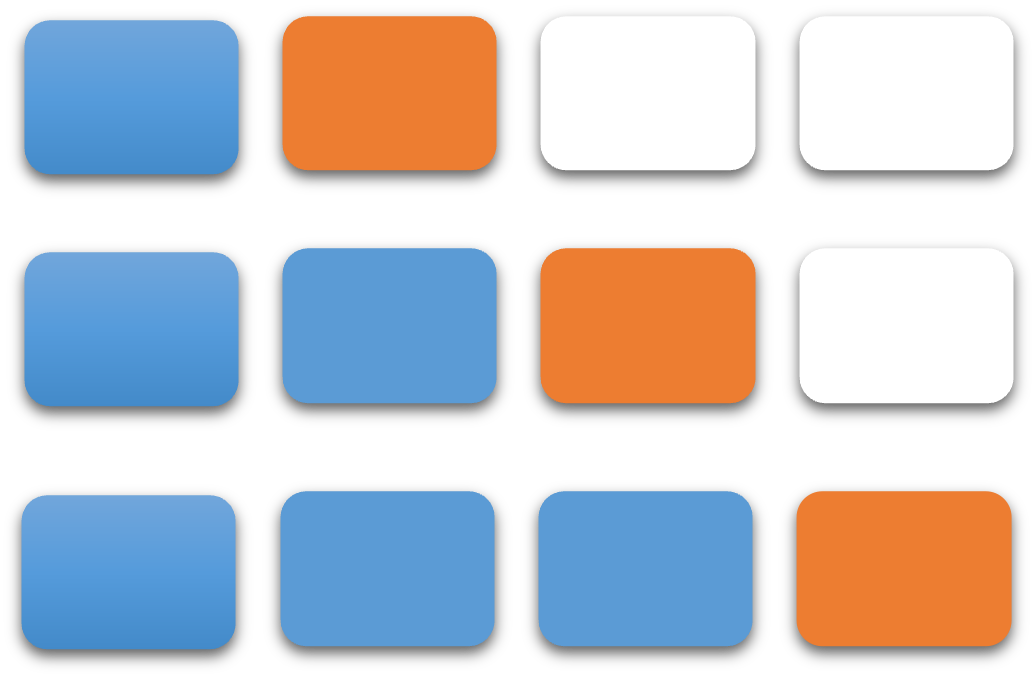}
\caption{Causal masking: consider a sequence with four tokens (chunks). We duplicate the sequence three times and progressively mask (represented in orange) one token within each duplicated sequence.}
\label{fig:res}
\end{figure}

\textbf{Causal Masking} We apply a novel causal masking scheme to the tokens generated by the embedding module. As illustrated in Fig.~\ref{fig:res}, we first duplicate the sequence of tokens. Starting from the second token, one token is masked and subsequent tokens are zeroed-out in each duplicated sequence. The masked token is replaced with a learnable token $\mathcal{M}$ of the same dimension. So, after causal masking, the input sequence to the GPT model is 
% $\{\mathcal{H}(D_1), \mathcal{M}, \mathbf{0}, \cdots, \mathbf{0} \},  \{\mathcal{H}(D_1), \mathcal{H}(D_2), \mathcal{M}, \mathbf{0}, \cdots, \mathbf{0}\},  \cdots, \{\mathcal{H}(D_1), \mathcal{H}(D_2), \mathcal{H}(D_3), \cdots, \mathcal{M} \}$
\begin{equation}
\centering
\begin{aligned}
&\{\mathcal{H}(D_1), \mathcal{M}, \mathbf{0}, \cdots, \mathbf{0}\}, \\
&\{\mathcal{H}(D_1), \mathcal{H}(D_2), \mathcal{M}, \mathbf{0}, \cdots, \mathbf{0}\}, \\
&\cdots, \\
&\{\mathcal{H}(D_1), \mathcal{H}(D_2), \mathcal{H}(D_3), \cdots, \mathcal{M}\}
\end{aligned}
\end{equation}
The pre-training of Neuro-GPT utilizes a self-supervised task, where the GPT model predicts every masked token in each sequence. We use a causal reconstruction loss defined in Eq.~\ref{eq:recon} as the self-supervised pre-training objective.
\begin{equation}
    \label{eq:recon}
    \begin{aligned}
        \mathcal{L} &= \frac{1}{N-1} \sum_{i=2}^N \| \hat{Y_i} - \mathcal{H}(D_i) \|_2 ^2 \\
        \text{where} \quad \hat{Y_i} &= \mathcal{G}[\mathcal{M} | \mathcal{H}(D_{i-1}), \mathcal{H}(D_{i-2}), \cdots, \mathcal{H}(D_{1})]
    \end{aligned}
\end{equation}
where $\mathcal{G}$ denotes the GPT model. We aggregate the reconstruction losses of masked tokens at each position. The predicted token $\hat{Y_i}$ produced by the GPT model is inferred based on the preceding tokens. By predicting the masked token separately from 1, 2, and 3 preceding tokens the model gains insight into the underlying temporal correlations in brain activity across different time scales. Thus the GPT model is potentially able to capture the dynamic evolution of brain activity more accurately.

\textbf{GPT Model} The GPT model employs a decoder-only transformer architecture consisting of a multi-layered stack of self-attention and feed-forward modules, enabling it to capture the global dependencies between tokens.
Unlike BERT~\cite{devlin2018bert}, which randomly masks some tokens in a sequence and the model predicts the masked tokens at random positions, GPT always predicts the next masked token given preceding tokens, also known as auto-regressive training~\cite{brown2020language}. This guarantees that the prediction of EEG embeddings considers the causal temporal relationship between tokens, thus improving our model of the underlying brain activity patterns.

\subsection{Pre-training}
\textbf{Pre-training Dataset}: 
The large-scale public dataset, Temple University Hospital (TUH) EEG corpus, is used as the pre-training dataset. TUH EEG corpus comprises a diverse archive of clinical EEG recordings from $14,987$ subjects with multiple sessions. The archive has over 40 different channel configurations and varying duration of recordings~\cite{obeid2016temple}. The sample frequency ranges from 250 - 1024 Hz, with the majority of recordings sampled at 250 Hz.

\textbf{Preprocessing}:
We preprocessed the TUH EEG dataset using the Brainstorm software~\cite{tadel2011brainstorm} in MATLAB (Mathworks, Inc.). Based on the channel labels, we selected 22 channels corresponding to the extended international 10-20 system (Fp1, Fp2, F7, F3, Fz, F4, F8, T1, T3, C3, Cz, C4, T4, T2, T5, P3, Pz, P4, T6, O1, Oz, O2). Channels with zero or missing signals throughout the recording sessions were marked as bad channels. The signals of the bad channels were interpolated by a weighted average of all neighboring channels with a maximal distance of 5cm between neighbors. EEG recordings were re-referenced to the average of 22 channels. We removed power line noise (60 Hz) using a notch filter and bandpass-filtered the data (0.5-100 Hz). All recordings were re-sampled to 250 Hz. We performed a DC offset correction and removed linear trends from the data. A z-transform was applied along the time dimension within each recording to normalize the data.

\textbf{Implementation Details}: 
During the pre-training phase, we simultaneously pre-train the entire Neuro-GPT model. After experimenting with various input configurations, we set the standard input as: 32 chunks, each with a length of 2 seconds and a $10\%$ (0.2 second) overlap. We randomly select a starting point for each EEG recording and then sample 32 contiguous chunks. If the total length of the EEG recording is shorter than the length to be sampled (57.8 seconds), we apply zero-padding to the end of the sequence. The attention weights are set to zero for the zero-padded part. In each training batch, one sampled sequence is considered as a single training sample. 
The EEG encoder consists of two convolutional layers followed by six self-attention layers, with an embedding dimension of 1,080. The first convolutional layer has a kernel size of $(1, 25)$, while the second has a kernel size of $(C, 1)$, with $C$ being the number of channels~\cite{song2023eeg}.

We employ the open-source GPT-2~\cite{radford2019language} model provided by Hugging Face~\cite{wolf2020huggingfaces}, which has an embedding dimension of $1024$. We specify $6$ transformer decoder layers in the GPT-2 model. A linear layer is added before the GPT-2 model to project the embedding dimension from 1080 to 1024. We pre-processed 20,000 EEG recordings from the TUH EEG dataset with a total duration of 5656 hours. 
We train the model on 19,000 EEG recordings for 135 epochs. The remaining 1000 EEG recordings were used as a hold-out validation set. 
% Following the training settings in~\cite{thomas2022self}, a linear learning rate decay schedule was adopted, including a warmup phase of $1\%$ of the total training steps. We applied the ADAM optimizer~\cite{kingma2014adam} with parameters ($\beta_1 = 0.9$, $\beta_2 = = 0.999$ , and $\epsilon = 1 \times 10^{-8}$),  gradient norm clipping at a threshold of 1.0 , and L2-regularization with a weight factor of 0.1.

\subsection{Downstream Fine-tuning}
\textbf{Downstream Dataset}: 
We define the downstream task as motor imagery classification, using the BCI Competition IV Dataset 2a provided by Graz University of Technology~\cite{brunner2008bci}. The BCI 2a dataset consists of nine subjects performing four motor imagery tasks: imagining left hand, right hand, feet, and tongue movement. Two sessions were collected on different days for each subject, using 22 Ag/AgCl electrodes at a sampling frequency of 250 Hz. Each recording has 72 trials per task, yielding a total of 288 trials. All trials from both sessions were used as training or testing samples - importantly, no subjects in the training data were included in the testing. Data was bandpass-filtered between 0.5 Hz and 100 Hz and normalized across time for each trial. We extract the sequence from $t=2s$ to $t=6s$ for each trial, which corresponds to the period when the cue and motor imagery tasks are performed. 

\textbf{Channel resampling}:
The downstream dataset has a different subset of 22 channel locations on the scalp from the pre-training dataset. To match the channel configuration between the two datasets, we resampled the downstream data to the pre-training dataset channel configuration using a $22\times 22$ transformation matrix. The transformation matrix was computed by solving the forward and the inverse problem for the source localization, mapping from one sensor configuration to the cerebral cortex and then back to the second configuration~\cite{mosher1999eeg, baillet2001electromagnetic}.

\textbf{Fine-tuning Details}: 
We fine-tune the pre-trained model on the BCI 2a dataset for the 4-class motor imagery classification task. To fully explore the potential of the foundation model, we designed three fine-tuning strategies:

\begin{enumerate}
    \item \textbf{Encoder-only}: Remove the GPT model and fine-tune the pre-trained EEG encoder only. (Note that in this case the model still benefits from including GPT in pre-training through the self-supervised training of the encoder in combination with the GPT model.) 
    % \item \textbf{GPT-only}: Drop the EEG encoder completely and only fine-tune the pre-trained GPT model.
    \item  \textbf{Encoder+GPT}: Fine-tune the entire Neuro-GPT model.
    \item \textbf{Linear}: Remove the GPT model, fix the EEG encoder and fine-tune only the linear head (3 linear layers).
\end{enumerate}
All strategies use the same pre-trained model and involve adding a linear head consisting of 3 linear layers to the end of the model for classification. For the Encoder+GPT strategy, we maintain the same number of chunks, the same chunk length, and the same overlapping ratio as used in the pre-training stage. Since only a 4-seconds sequence is extracted from each EEG recording in the BCI 2a dataset, we apply zero-padding to the end of the sequence. 
% For GPT-only, the inputs to the GPT model are chunks of raw EEG signals flattened across channels.
% In this case, only three chunks contain actual EEG signals. However, owing to the causal masking applied during pre-training, the foundation model is capable of capturing latent patterns from any subset of chunks in the sequence. The fine-tuned foundation model will be aware of the zero signals in the sequence and focus on learning meaningful information from the first three chunks. 
In the Encoder-only strategy, we feed the model with two non-overlapping 2-second chunks, and no zero-padding is applied. For the Linear strategy, all the pre-trained parameters from the EEG encoder are frozen during fine-tuning. We only fine-tune the linear head, which takes the output features of the EEG encoder as input. No masking is applied during fine-tuning.

\section{Experiments and Results}
\textbf{Fine-tuning Classification Performance}: 
Unlike previous studies which only focused on within-subject classification~\cite{zhang2021eeg, song2023eeg}, we performed leave-one-subject-out cross-validation, which is more challenging due to the high inter-subject variance. We compute the average classification accuracy across subjects. To explore the benefits of applying a pre-trained foundation model, we compare the classification performance of a model trained from scratch (w/o pre-training) to that of the same model fine-tuned on the pre-trained foundation model (w/ pre-training). In addition, we compare the proposed Neuro-GPT with BENDR~\cite{kostas2021bendr}, a BERT-inspired transformer model trained on TUH EEG data using contrastive self-supervised learning and then fine-tuned on the BCI classification data. As shown in Table~\ref{tab:res}, Neuro-GPT significantly improved the classification performance compared with the best performance of BENDR, and outperforms other methods for motor imagery classification using leave-one-subject-out cross-validation.
\begin{table}[!b]
\resizebox{\linewidth}{!}{%
\centering
\begin{tabular}{c|c c}
\toprule[1.5pt]
Method & w/o Pre-train & w/ Pre-train \\
\hline
 Linear & $0.398 \pm 0.054$ & $0.443 \pm 0.051$ \\
 % GPT-only & $0.497 \pm 0.061$ & $0.507 \pm 0.063$ \\
 Encoder-only & $0.606 \pm 0.098$ & $\mathbf{0.645 \pm 0.104}$ \\
Encoder+GPT & $0.596 \pm 0.090$& $0.586 \pm 0.098$ \\
\hline
BENDR~\cite{kostas2021bendr} & / & $0.426$ \\
SVM~\cite{oikonomou2017comparison} & $0.361 \pm 0.082$ & / \\
EEGNet~\cite{lawhern2018eegnet} & $0.513 \pm 0.052$ & / \\
CTCNN~\cite{schirrmeister2017deep} & $0.477 \pm 0.151$ & / \\
CCNN~\cite{amin2019deep} & $0.553 \pm 0.101$ & / \\
NG-CRAM~\cite{zhang2020motor}  & $0.601 \pm 0.102$ & / \\
\bottomrule[1.5pt]
\end{tabular}
}
\caption{A comparison of means and stds of four-class classification accuracy among different methods. The first three rows are three fine-tuning strategies of Neuro-GPT, accuracies reported in other work are shown in the bottom rows.}
\label{tab:res}
\end{table}

The performance of models with pre-training surpassed that of models without pre-training for both Linear and Encoder-only fine-tuning strategies, highlighting that applying a foundation model to a downstream task can lead to effective feature learning and, consequently, improved performance. Among the fine-tuning strategies, Encoder-only achieved the best performance, indicating that the encoder learned expressive and generalizable features during pre-training, thus facilitating the learning of distinguishable features for downstream tasks. 
% This outcome verifies our assumption that the EEG encoder should encode intrinsic spatio-temporal features of EEG, which are crucial to the classification of motor imagery tasks.
The Encoder+GPT yielded worse performance, possibly because the GPT model only serves as an auxiliary component to assist the EEG encoder in encoding meaningful features from raw EEG data. 
% The GPT model learns features related to inferring the next token, which may be challenging to transfer to classification tasks. 
The GPT model has more trainable parameters than the encoder. Fine-tuning a large model on a small data-set can lead to over-fitting. 
% However, 
% distinguishing between different motor imagery tasks requires spatial EEG signals from multiple brain regions~\cite{song2023eeg}. So relying only on temporal information is insufficient for accurate classification. 
To examine whether the features learned by the foundation model are linearly separable, we input the features generated by the EEG encoder to the linear head for classification. The classification accuracy achieved by fine-tuning only the linear head is $0.443$ vs. $0.398$ with out pre-training, indicating that the EEG encoder can encode meaningful features through pre-training.

\textbf{Hyper-parameter Evaluation in Pre-training}:
To explore the optimal input configurations for the foundation model during pre-training, we conducted experiments with varying numbers of chunks (4, 8, 16, 32), chunk lengths (1s, 2s, 4s), and overlapping ratios ($10\%$, $50\%$). Different model architectures were also investigated. Key findings include:
\begin{itemize}
    \item Chunks with a 1-second length are more straightforward to predict (as embedded tokens) but led to poorer downstream performance.
    \item Chunks with longer lengths are more challenging to predict but enhance downstream performance.
    \item Increasing the number of chunks is beneficial. Training with 32 and 16 chunks yielded better downstream results than training with 8 or 4 chunks.
    \item Increasing the overlapping ratio to $50\%$ improved reconstruction, but degraded the downstream performance.
    \item Increasing the embedding dimension of GPT-2 model ($768\rightarrow 1024$) improved downstream performance.
    \item Reducing the number of self-attention layers in the encoder ($6\rightarrow 4, 2$) degraded downstream performance.
    \item Adding more GPT decoder layers ($6\rightarrow 8, 10$) did not improve downstream performance.
\end{itemize}
% Considering both computational efficiency and performance gains, we set the standard input configuration as: 32 chunks, each with a length of 2 seconds and a $10\%$ overlap.

\section{Discussion}
We have demonstrated that pre-training a foundation model on a large-scale EEG dataset boosts downstream task performance. Through exploring different fine-tuning strategies, we discovered that the pre-trained EEG encoder captures inherent and fundamental features of EEG that are generalizable across datasets, leading to significant improvements in classification performance. 

\section{Acknowledgment}
This project is sponsored in part by the NIH under grant R01 EB026299 and in part by the Defense Advanced Research Projects Agency (DARPA) under cooperative agreement No. N660012324006. The content of the information does not necessarily reflect the position or the policy of the Government, and no official endorsement should be inferred.

\bibliographystyle{IEEEbib}
\bibliography{refs2}

\end{document}